\documentclass{ecai} 

\usepackage{latexsym}
\usepackage{amssymb}
\usepackage{amsmath}
\usepackage{amsthm}
\usepackage{booktabs}
\usepackage{enumitem}
\usepackage{graphicx}
\usepackage{color}
\usepackage{xcolor}
\usepackage{bm}
\usepackage[hidelinks]{hyperref}
\hypersetup{
    colorlinks=true,
    citecolor=blue,
    linkcolor=blue,     
    urlcolor=blue,
    pdftitle={Overleaf Example},
    pdfpagemode=FullScreen,
    }

\usepackage{nicefrac}       
\usepackage{microtype}      
\usepackage{array,colortbl}
\usepackage{algorithm,algorithmicx,algpseudocode}
\usepackage[capitalise]{cleveref}
\usepackage{graphbox}
\usepackage{placeins}
\usepackage{wrapfig}
\usepackage{etoolbox}
\usepackage{subcaption}
\usepackage{xcolor}
\usepackage{multirow}

\definecolor{ourmethod}{gray}{0.93}

\newcommand{\bx}{\mathbf{x}}
\newcommand{\bz}{\mathbf{z}}
\newcommand{\bu}{\mathbf{u}}
\newcommand{\by}{\mathbf{y}}
\newcommand{\bA}{\mathbf{A}}

\newcommand{\zc}{\bz_{\text{causal}}}

\newcommand{\BibTeX}{B\kern-.05em{\sc i\kern-.025em b}\kern-.08em\TeX}

\begin{document}


\begin{frontmatter}


\paperid{1635} 


\title{Causal Diffusion Autoencoders: Toward Counterfactual Generation via Diffusion Probabilistic Models}


\author[1]{\fnms{Aneesh}~\snm{Komanduri}\thanks{Corresponding Author. Email: akomandu@uark.edu.}}
\author[2]{\fnms{Chen}~\snm{Zhao}}
\author[3]{\fnms{Feng}~\snm{Chen}}
\author[1]{\fnms{Xintao}~\snm{Wu}}
\address[1]{University of Arkansas}
\address[2]{Baylor University}
\address[3]{University of Texas at Dallas}


\begin{abstract}
    Diffusion probabilistic models (DPMs) have become the state-of-the-art in high-quality image generation. However, DPMs have an arbitrary noisy latent space with no interpretable or controllable semantics. Although there has been significant research effort to improve image sample quality, there is little work on representation-controlled generation using diffusion models. Specifically, causal modeling and controllable counterfactual generation using DPMs is an underexplored area. In this work, we propose CausalDiffAE, a diffusion-based causal representation learning framework to enable counterfactual generation according to a specified causal model. Our key idea is to use an encoder to extract high-level semantically meaningful causal variables from high-dimensional data and model stochastic variation using reverse diffusion. We propose a causal encoding mechanism that maps high-dimensional data to causally related latent factors and parameterize the causal mechanisms among latent factors using neural networks. To enforce the disentanglement of causal variables, we formulate a variational objective and leverage auxiliary label information in a prior to regularize the latent space. We propose a DDIM-based counterfactual generation procedure subject to do-interventions. Finally, to address the limited label supervision scenario, we also study the application of CausalDiffAE when a part of the training data is unlabeled, which also enables granular control over the strength of interventions in generating counterfactuals during inference. We empirically show that CausalDiffAE learns a disentangled latent space and is capable of generating high-quality counterfactual images. 
\end{abstract}

\end{frontmatter}


\section{Introduction}

Diffusion probabilistic models (DPMs) \cite{pmlr-v37-sohl-dickstein15, ho_ddpm, improved_diffusion, song2021denoising, song2021scorebased} are a class of likelihood-based generative models that have achieved remarkable successes in the generation of high-resolution images with many large-scale implementations such as DALLE-2 \citep{ramesh2022hierarchical}, Stable Diffusion \citep{Rombach_2022_CVPR}, and Imagen \citep{saharia2022photorealistic}. Thus, there has been 
great interest in evaluating the capabilities of diffusion models. Two of the most promising approaches are formulated as discrete-time \citep{ho_ddpm} and continuous-time \citep{song2021scorebased} step-wise perturbations of the data distribution. A model is then trained to estimate the reverse process which transforms noisy samples to samples from the underlying data distribution. Representation learning has been an integral component of generative models such as GANs \citep{NIPS2014_5ca3e9b1} and VAEs \citep{kingma2013} for extracting robust and interpretable features from complex data \citep{scholkopf_toward_2021, bengio2014representation, radford2021learning}. Recently, a thrust of research has focused on whether DPMs can be used to extract a semantically meaningful and decodable representation that increases the quality of and control over generated images \citep{pandey2022diffusevae, preechakul2021diffusion}. However, there has been no work in modeling causal relations among the semantic latent codes to learn causal representations and enable \textit{counterfactual generation} at inference time in DPMs. Generating high-quality counterfactual images is critical for domains such as healthcare and medicine \citep{LIU2022102516, sanchez2022causal}. The ability to generate accurate counterfactual data from a causal graph obtained from domain knowledge can significantly cut the cost of data collection. Furthermore, reasoning about hypothetical scenarios unseen in the training distribution can be quite insightful for gauging the interactions among causal variables in complex systems. Given a causal graph of a system, we study the capability of DPMs as causal representation learners and evaluate their ability to generate counterfactuals upon interventions on causal variables.

Intuitively, we can think about the DPM as an encoder-decoder framework. The encoding maps an input image $\bx_0$ to a spatial latent variable $\bx_T$ through a series of Gaussian noise perturbations. However, $\bx_T$ can be interpreted as a noise representation that lacks high-level semantics \citep{preechakul2021diffusion}. Recently, Preechakul et al \citep{preechakul2021diffusion} proposed a diffusion-based autoencoder (DiffAE) to extract a high-level semantic representation alongside the stochastic low-level representation $\bx_T$ for decodable representation learning. Learning such a semantic representation also enables interpolation in the latent space for controllable generation and has been shown to improve image generation quality. Mittal et al \citep{pmlr-v202-mittal23a} built on this framework and introduced a diffusion-based representation learning (DRL) objective that instead learns time-conditioned representations throughout the diffusion process. However, both these approaches learn arbitrary representations and do not focus on disentanglement, a key property of interpretable representations. Disentangled representations enable precise control of generative factors in isolation. When considering causal systems, disentanglement is important for performing isolated interventions.

In this paper, we focus on learning disentangled causal representations, where the high-level semantic factors are causally related. To the best of our knowledge, we are the first to explore representation-based counterfactual image generation using diffusion probabilistic models. We propose \textbf{CausalDiffAE}, a learning framework for causal representation learning and controllable counterfactual generation in DPMs. Our key idea is to learn a causal representation via a learnable stochastic encoder and model the relations among latents via causal mechanisms parameterized by neural networks. We formulate a variational objective with a label alignment prior to enforce disentanglement of the learned causal factors. We then utilize a conditional denoising diffusion implicit model (DDIM) \citep{song2021denoising} for decoding and modeling the stochastic variations. Intuitively, the causal representation encodes compact information that is \textit{causally relevant} for image decoding in reverse diffusion. Furthermore, the modeling of causal relations in the latent space enables the generation of counterfactuals upon interventions on learned causal variables. We propose a DDIM variant for counterfactual generation subject to $\textbf{do}(\cdot)$ interventions \citep{Pearl09}. In an effort to improve the practicality and interpretability of the model, we propose an extension to CausalDiffAE that utilizes weaker supervision. In the scenario where labeled data is limited, we jointly train an unconditional and representation-conditioned diffusion model on the unlabeled and labeled partitions, respectively. This approach significantly reduces the number of labeled samples required for training and enables granular control over the strength of interventions and the quality of generated counterfactuals.

\section{Related Work}
Recent work in causal generative modeling has focused on either learning causal representations or controllable counterfactual generation \citep{komanduri2024identifiable}. Yang et al proposed CausalVAE \citep{DBLP:conf/cvpr/YangLCSHW21}, a causal representation learning framework that models latent causal variables by a linear SCM. Kocaoglu et al \citep{kocaoglu2017causalgan} proposed CausalGAN, an extension of the GAN to model causal variables for sampling from interventional distributions. Diffusion and score-based generative models \citep{ho_ddpm, song2021scorebased} have shown impressive results in class-conditional generation either through classifier-based \citep{dhariwal2021diffusion} or classifier-free \citep{ho2021classifierfree} paradigms. Recently, there has been an interest in exploring the capacity of diffusion models as representation learners. For instance, Mittal et al \citep{pmlr-v202-mittal23a} and Preechakul et al \citep{preechakul2021diffusion} considered diffusion-based representation learning objectives. Mamaghan et al \citep{mamaghan2023diffusion} explored representation learning from a score-based perspective given access to data in the form of counterfactual pairs. However, this work does not focus on counterfactual generation. Another related area of research is counterfactual explanations \citep{augustin2022diffusion}, which focuses on post-hoc methods to generate realistic counterfactuals, but not in the strictly causal sense. Our work focuses on diffusion-based \textit{representation learning} and is most closely related to DiffAE \citep{preechakul2021diffusion} and DRL \citep{pmlr-v202-mittal23a}, which aim to learn semantically meaningful representations. However, the key distinction is that we learn causal representations to enable counterfactual generation. Our proposed framework extends CausalVAE to diffusion-based models and under a weaker supervision paradigm.


\section{Background}

\subsection{Structural Causal Model}
A structural causal model (SCM) is formally defined by a tuple $\mathcal{M} = \langle \mathcal{Z}, \mathcal{U}, F\rangle$, where $\mathcal{Z}$ is the domain of the set of $n$ endogenous causal variables $\bz=\{z_1, \dots, z_n\}$, $\mathcal{U}$ is the domain of the set of $n$ exogenous noise variables $\bu=\{u_1, \dots, u_n\}$, which is learned as an intermediate latent variable, and $F = \{f_1, \dots, f_n\}$ is a collection of $n$ independent causal mechanisms of the form
\begin{equation}
    z_i = f_i(u_i, z_{\textbf{pa}_i})
\label{eq:prelim_scm}
\end{equation}
where $\forall i$, $f_i: \mathcal{U}_i \times \prod_{j\in \textbf{pa}_i} \mathcal{Z}_j \to \mathcal{Z}_i$ are \textbf{causal mechanisms} that determine each causal variable as a function of the parents and noise, $z_{\textbf{pa}_i}$ are the parents of causal variable $z_i$; and a probability measure $p_{\mathcal{U}}(\bu) = p_{\mathcal{U}_1}(u_1)p_{\mathcal{U}_2}(u_2)\dots p_{\mathcal{U}_n}(u_n)$, which admits a product distribution. For the purposes of this work, we assume a causally sufficient setting (no hidden confounding), no SCM misspecification, and faithfulness is satisfied.

\begin{figure*}
    \centering
\includegraphics[width=0.95\textwidth]{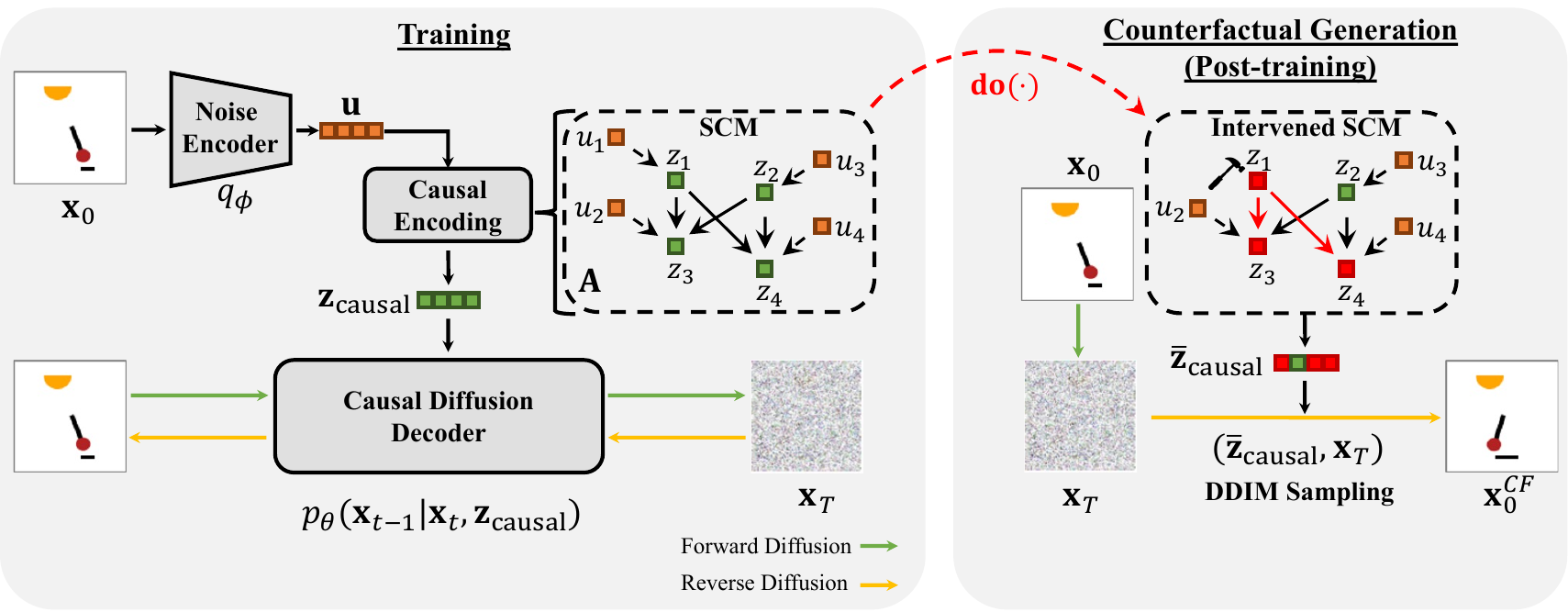}
\vspace{0.1cm}
    \caption{CausalDiffAE Framework. The left side details the training process of CausalDiffAE by encoding to causal representation $\zc$ and using a conditional DDIM decoder conditioned on $\zc$ and $\bx_T$ for image reconstruction. The right side shows the DDIM-based counterfactual generation procedure using a trained CausalDiffAE model.}
    \vspace{0.5cm}
    \label{fig:architecture}
\end{figure*}

\subsection{Diffusion Probabilistic Models}
Diffusion Probabilistic Models (DPMs) \cite{ho_ddpm, improved_diffusion} have shown impressive results in image generation tasks, even beating out GANs in many cases \cite{dhariwal2021diffusion}. The idea of the denoising diffusion probabilistic model (DDPM) \citep{ho_ddpm} is to define a Markov chain of diffusion steps to slowly destroy the structure in a data distribution through a forward diffusion process by adding noise \cite{ho_ddpm} and learn a reverse diffusion process that restores the structure of the data. Some proposed methods, such as denoising diffusion implicit model (DDIM) \cite{song2021denoising}, break the Markov assumption to speed up the sampling in the diffusion process by carrying out a deterministic encoding of the noise. 

\textbf{Forward Diffusion.} Given some input data sampled from a distribution $\bx_0 \sim q(\bx)$, the forward diffusion process is defined by adding small amounts of Gaussian noise to the sample in $T$ steps thereby producing noisy samples $\bx_1, \dots, \bx_T$. The distribution of the noisy sample at time step $t$ is defined as a conditional distribution as follows:

\begin{equation}
    q(\bx_t | \bx_{t-1}) = \mathcal{N}(\bx_t; \sqrt{1-\beta_t} \bx_{t-1}, \beta_t \mathbf{I})
    \label{eq:analytical_forward}
\end{equation}

where $\beta_t\in(0, 1)$ is a variance parameter that controls the step size of noise. As $t\to \infty$, the input sample $\bx_0$ loses its distinguishable features. In the end, when $t=T$, $\bx_T$ follows an isotropic Gaussian. From Eq (\ref{eq:analytical_forward}), we can then define a closed-form tractable posterior over all time steps factorized as follows:

\begin{equation}
    q(\bx_{1:T} | \bx_0) = \prod_{t=1}^T q(\bx_t | \bx_{t-1})
    \label{eq:joint_analytical_forward}
\end{equation}

Now, $\bx_t$ can be sampled at any arbitrary time step $t$ using the reparameterization trick. Let $\alpha_t = \prod_{i=1}^t 1-\beta_i$:

\begin{equation}
    q(\bx_t | \bx_0) = \mathcal{N}(\bx_t; \sqrt{\alpha_t}\bx_0, (1-\alpha_t)\mathbf{I})
\end{equation}

\textbf{Reverse Diffusion.} In the reverse process, to sample from $q(\bx_{t-1}|\bx_t)$, the goal is to recreate the true sample $\bx_0$ from a Gaussian noise input $\bx_T\sim \mathcal{N}(\mathbf{0}, \mathbf{I})$. Unlike the forward diffusion, $q(\bx_{t-1}|\bx_t)$ is not analytically tractable and thus requires learning a model $p_{\theta}$ to approximate the conditional distributions as follows:

\begin{equation}
\begin{split}
    p_{\theta}(\bx_{0:T}) &= p(\bx_T) \prod_{t=1}^T p_{\theta}(\bx_{t-1}|\bx_t) \\
    p_{\theta}(\bx_{t-1}|\bx_t) &= \mathcal{N}(\bx_{t-1}; \mu_{\theta}(\bx_t, t), \Sigma_{\theta}(\bx_t, t))
\end{split}
\end{equation}

where $\mu_{\theta}$ and $\Sigma_{\theta}$ are learned via neural networks. It turns out that conditioning on the input $\bx_0$ yields a tractable reverse conditional probability 

\begin{equation}
    q(\bx_{t-1}|\bx_t, \bx_0) = \mathcal{N}(\bx_{t-1}; \tilde{\mu}(\bx_t, \bx_0), \tilde{\beta}_t \mathbf{I})
\end{equation}

where $\tilde{\mu}$ and $\tilde{\beta}_t$ are the true mean and variance. The learning objective is then formulated as a simplified objective of the ELBO via reparameterization to minimize the following mean squared error loss

\begin{equation}
    \mathcal{L}_{\text{simple}} = \sum_{t=1}^T \mathbb{E}_{\bx_0, \epsilon_t}\Big[\|\epsilon_t - \epsilon_{\theta}(\bx_t, t)\|^2_2\Big]
\end{equation}

where $\epsilon_t \sim \mathcal{N}(\mathbf{0}, \mathbf{I})$ is the noise that takes an analytical form via a reparameterization from $\bx_0$, as shown in \cite{ho_ddpm}. 

DPMs produce latent variables $\bx_{1:T}$ through the forward process. However, these variables are stochastic \citep{preechakul2021diffusion}. Song et al. proposed a DPM called Denoising Diffusion Implicit model (DDIM), which enables a deterministic process as follows:

\begin{equation}
    \bx_{t-1} = \sqrt{\alpha_{t-1}}\Big(\frac{\bx_t - \sqrt{1-\alpha_t} \epsilon_{\theta}(\bx_t, t)}{\sqrt{\alpha_t}}\Big) + \sqrt{1 - \alpha_{t-1}}\epsilon_{\theta}(\bx_t, t)
\end{equation} 

with the following deterministic decoding process

\begin{equation}
    q(\bx_{t-1} | \bx_t, \bx_0) = \mathcal{N}\Big(\sqrt{\alpha_{t-1}}\bx_0 + \sqrt{1 - \alpha_{t-1}} \frac{\bx_t - \sqrt{\alpha_t}\bx_0}{\sqrt{1-\alpha_t}}, \mathbf{0}\Big)
\end{equation}

which keeps the DDPM marginal distribution $q(\bx_t | \bx_0) = \mathcal{N}(\sqrt{\alpha_{t-1}}\bx_0, (1-\alpha_t)\mathbf{I})$. It turns out that this formulation shares the same objective and solution of DDPM and only differs in the sampling procedure. Thus, we can \textit{deterministically} obtain the noise map $\bx_T$ corresponding to a given image $\bx_0$.

\section{Causal Diffusion Autoencoders}
Existing diffusion-based controllable generation methods neglect the scenario where generative factors are causally related and do not support counterfactual generation. To tackle this issue, we propose CausalDiffAE, a diffusion-based causal representation learning framework to enable counterfactual generation. Firstly, we define a latent SCM to describe the semantic causal representation as a function of learned noise encodings. In the case of diffusion autoencoders \citep{preechakul2021diffusion}, the semantic latent representation $\bz_{\text{sem}}$ captures high-level semantic information, and $\bx_T$ captures low-level stochastic information. In our formulation, we learn a causal representation $\bz_{\text{causal}}$ which captures \textit{causally relevant} information. Together, the two latent variables $(\bz_{\text{causal}}, \bx_T)$ capture all the detailed causal semantics and stochasticity in the image. Secondly, given a trained CausalDiffAE model, we propose a counterfactual generation algorithm that utilizes $\textbf{do}(\cdot)$ interventions and the DDIM sampling algorithm. The overall framework of CausalDiffAE is shown in Figure \ref{fig:architecture}.

\subsection{Causal Encoding}
Let $\bx_0 \in \mathbb{R}^d$ be the observed input image. We carry out the forward diffusion process until we have a set of $T$ perturbed samples $\{\bx_1, \bx_2, \dots, \bx_{T}\}$, each at a different noise scale. Suppose there are $n$ abstract causal variables that describe the high-level semantics of the observed image. To learn a meaningful representation, we propose to encode the input image $\bx_0$ to a low-dimensional noise encoding $\bu \in \mathbb{R}^n$. We then map the noise encoding to latent causal factors $\bz_{\text{causal}}\in \mathbb{R}^n$ corresponding to the abstract causal variables. In this formulation, each noise term $u_i$ is the exogenous noise term for causal variable $z_i$ in the SCM. Let $\mathbf{A}$ be the adjacency matrix encoding the causal graph among the underlying factors where $A_{ji}$ implies $z_j$ is a cause of $z_i$. Then, we parameterize the mechanisms among causal variables as follows

\begin{equation}
    z_{i} = f_i(z_{\textbf{pa}_i}, u_i)
\end{equation}

where $f_i$ is the causal mechanism generating causal variable $z_i$ as a function of its parents and exogenous noise term and $z_{\textbf{pa}_i}$ denotes the causal parents of factor $z_i$. In practice, we can implement $f_i$ as a post-nonlinear additive noise model such that

\begin{equation}
\begin{split}
    \bz &= (I - \bA^T)^{-1}\bu \\
    z_i &= f_i(\bA_i \odot \bz; \nu_i) + u_i
\end{split}
\end{equation}

where $\nu_i$ are the parameters of the neural network parameterizing each mechanism, $\odot$ is the elementwise product, and $\zc = \{z_1, \dots, z_n\}$. This module captures the causal relations between latent variables using neural structural causal models. For the purposes of this work, we assume that the causal graph is known since we focus on counterfactual generation. However, a more end-to-end framework may include a causal discovery component. See Appendix \ref{sec:limitations} for a more detailed discussion.

\subsection{Generative Model}

Let $\bx_0$ denote the high-dimensional input image and $\by\in \mathbb{R}^n$ denote an auxiliary weak supervision signal. Then, the CausalDiffAE generative process can be factorized as follows:

\begin{equation}
    p(\bx_{0:T}, \bu, \bz_{\text{causal}} | \by) = p_{\theta}(\bx_{0:T} | \bu, \zc, \by)p(\bu, \zc|\by)
\end{equation}

where $\theta$ are the parameters of the reverse process of the causal diffusion decoder (will discuss in Section \ref{sec:diff_dec}), $p(\bu, \zc | \by) = p(\bu) p(\zc | \by)$, $p(\bu) = \mathcal{N}(\mathbf{0}, \mathbf{I})$, and $p(\zc | \by)$ is the alignment prior defined in Eq. (\ref{eq:alignment_prior}). The joint posterior distribution $p(\bx_{1:T}, \bu, \zc | \bx_0, \by)$ is intractable, so we approximate it using a variational distribution $q(\bx_{1:T}, \bu, \zc | \bx_0, \by)$ which can be factorized into the following conditional distributions

\begin{equation}
\begin{split}
    q(\bx_{1:T}, \bu, \zc | \bx_0, \by) = q_{\phi}&(\zc, \bu | \bx_0, \by) \\ &q(\bx_{1:T}|\bu, \zc, \bx_0)
\end{split}
\label{eq:posterior}
\end{equation}

where $\phi$ are the parameters of the variational encoder network parameterizing the joint distribution over the noise $\bu$ and causal factors $\zc$. We can remove the dependence on $\by$ for the second conditional term in the decomposition of Eq. (\ref{eq:posterior}) since $\bx_{1:T}$ is independent of the auxiliary label $\by$. We note that $q_{\phi}(\zc, \bu | \bx_0, \by)$ can be factorized as $q_{\phi}(\zc | \bx_0, \by)q_{\phi}(\bu | \bx_0)$ since $\bu$ and $\zc$ have a one-to-one correspondence.

\begin{algorithm}[t]
  \caption{CausalDiffAE Training} \label{alg:training}
   \textbf{Input:} (image, label) pairs $(\bx_0, \by)$ \\
  \textbf{Output:} learned parameters $\{\theta, \phi\}$
  \begin{algorithmic}[1]
    \Repeat
      \State $\bx_0 \sim q(\bx_0)$
      \State $\bu \sim q_{\phi}(\bu | \bx_0)$    \Comment{Noise encoding}
      \State $\bz_{\text{causal}} = \{f_i(u_i, z_{\textbf{pa}_i};\nu_i)\}_{i=1}^n$ \Comment{Causal encoding}
      \State $t \sim \mathcal{U}(\{1, \dots, T\})$ \Comment{Sample timestep}
      \State $\epsilon_t \sim\mathcal{N}(\mathbf{0}, \mathbf{I})$
      \State $\bx_t = \sqrt{\alpha_t} \bx_0 + \sqrt{1-\alpha_t}\epsilon_t$ \Comment{Corrupt data to sampled time}
      \State Take gradient step on $\nabla_{\theta, \phi}\mathcal{L}_{\text{CausalDiffAE}}$ 
    \Until{convergence}
  \end{algorithmic}
\end{algorithm}

\subsection{Causal Diffusion Decoder}
\label{sec:diff_dec}
 We use a conditional DDIM decoder that takes as input the pair of latent variables $(\bz_{\text{causal}}, \bx_T)$ to generate the output image. We approximate the inference distribution $q(\bx_{t-1} | \bx_t, \bx_0)$ by parameterizing the probabilistic decoder via a conditional DDIM $p_{\theta}(\bx_{t-1} | \bx_t, \bz_{\text{causal}})$. With DDIM, the forward process becomes completely deterministic except for $t=1$. Similar to \citep{preechakul2021diffusion}, we define the joint distribution of the reverse generative process as follows:
 
\begin{equation}
    p_{\theta}(\bx_{0:T}|\bz_{\text{causal}}) = p(\bx_T) \prod_{t=1}^Tp_{\theta}(\bx_{t-1}|\bx_t, \bz_{\text{causal}})
    \label{eq:causal_diffusion_decoder}
\end{equation}

\begin{equation}
    p_{\theta}(\bx_{t-1}|\bx_t, \bz_{\text{causal}}) = \begin{cases}
        \mathcal{N}(\mathbf{f}_{\theta}(\bx_1, 1, \zc), \mathbf{0}) & \text{if $t=1$} \\
        q(\bx_{t-1}|\bx_t, \mathbf{f}_{\theta}(\bx_t, t, \bz_{\text{causal}})) & \text{otherwise}
    \end{cases}
\end{equation}

where $\mathbf{f}_{\theta}$ is parameterized by a noise prediction network $\epsilon_{\theta}$ (i.e., UNet \citep{dhariwal2021diffusion}) as follows:

\begin{equation}
    \mathbf{f}_{\theta}(\bx_t, t, \zc) = \frac{1}{\sqrt{\alpha_t}}(\bx_t - \sqrt{1-\alpha_t} \epsilon_{\theta}(\bx_t, t, \zc))
\end{equation}

Note that in Eq. (\ref{eq:causal_diffusion_decoder}), $\bu$ is omitted since $\zc$ already captures all the information about the noise. By leveraging the reparameterization trick, we can optimize the following mean squared error between noise terms 

\begin{equation}
    \mathcal{L}_{\text{simple}} = \sum_{t=1}^T \mathbb{E}_{\bx_0, \epsilon_t} \Big[ \|\epsilon_{\theta}(\bx_t, t, \bz_{\text{causal}}) - \epsilon_t \|_2^2\Big]
\end{equation}

where $\epsilon_t \sim \mathcal{N}(\mathbf{0}, \mathbf{I})$ and $\bx_t = \sqrt{\alpha_t} \bx_0 + \sqrt{1-\alpha_t}\epsilon_t$.

\subsection{Learning Objective}
To ensure the causal representation is disentangled, we incorporate label information $\by \in \mathbb{R}^n$ as a prior in the variational objective to aid in learning semantic factors and for identifiability guarantees \cite{pmlr-v108-khemakhem20a}. We define the following joint loss objective:

\begin{equation}
\begin{split}
    \mathcal{L}_{\text{CausalDiffAE}} &= \mathcal{L}_{\text{simple}} \\& + \gamma \Big\{\mathcal{D}_{KL}(q_{\phi}(\bz_{\text{causal}} | \bx_0, \by) \| p(\bz_{\text{causal}} | \by)) \\& + \mathcal{D}_{KL}(q_{\phi}(\bu | \bx_0) \| \mathcal{N}(\mathbf{0}, \mathbf{I}))\Big\}
\end{split}
\end{equation}

where $\gamma$ is a regularization hyperparameter similar to the bottleneck parameter in $\beta$-VAEs \citep{higgins2017betavae}, and the alignment prior over latent variables is defined as the following exponential family distribution

\begin{equation}
    p(\bz_{\text{causal}}|\by) = \prod_{i=1}^n p(z_i | y_i) = \prod_{i=1}^n \mathcal{N}(z_i; \mu_{\nu}(y_i), \sigma^2_{\nu}(y_i)\mathbf{I})
    \label{eq:alignment_prior}
\end{equation}

where $\mu_{\nu}$ and $\sigma^2_{\nu}$ are functions that estimate the mean and variance of the Gaussian, respectively. Intuitively, this prior ensures that the learned factors are one-to-one mapped to an indicator of the underlying ground truth factors. DiffAE requires training a latent DDIM in the latent space of the pre-trained autoencoder to enable sampling of latent semantic representation. However, CausalDiffAE is formulated as a variational objective with a stochastic encoder. Thus, we can sample the representation from the defined prior directly without having to train a separate diffusion model in the latent space. The training procedure for CausalDiffAE is outlined in Algorithm \ref{alg:training}. See Appendix \ref{sec:elbo} for a derivation of the ELBO. For a detailed discussion on the connection of our diffusion objective to score-based generative models \citep{song2021scorebased}, see Appendix \ref{sec:score_connection}.

\begin{algorithm}[t]
  \caption{CausalDiffAE Counterfactual Generation} \label{alg:generation}
  \small
   \textbf{Input:} Factual sample $\bx_0$, intervention target set $\mathcal{I}$ with intervention values $c$, noise predictor $\epsilon_{\theta}$, encoder $\phi$ \\
  \textbf{Output:} Counterfactual sample $\bx_0^{CF}$
  \begin{algorithmic}[1]
    \vspace{.04in}
    \State $\bu \sim  q_{\phi}(\bu | \bx_0)$ \Comment{Noise encoding}
      \For{$i=1$ to $n$} \Comment{in topological order}
      \If{$i\in \mathcal{I}$}
        \State $z_i = c_i$
      \Else 
        \State $z_{i} = f_i(u_i, z_{\textbf{pa}_i})$
      \EndIf
      \EndFor
    \State $\bar{\mathbf{z}}_{\text{causal}} = \{z_1, \dots, z_n\}$ \Comment{Intervened representation}
    \State $\bx_T \sim \mathcal{N}(\sqrt{\alpha_T}\bx_0, (1-\alpha_T)\mathbf{I})$
    \State $\bx_T^{CF} = \bx_T$
    \For{$t=T, \dotsc, 1$} \Comment{DDIM sampling}
      \State $\bx_{t-1}^{CF} = \sqrt{\alpha_{t-1}}\Big(\frac{\bx_t^{CF} - \sqrt{1-\alpha_t}\epsilon_{\theta}(\bx_t^{CF}, t, \mathbf{z}_{\text{causal}})}{\sqrt{\alpha_t}}\Big)$ \\\hspace{3cm} $+ \sqrt{1 - \alpha_{t-1}}\epsilon_{\theta}(\bx_t^{CF}, t, \mathbf{z}_{\text{causal}})$
    \EndFor
    \State \textbf{return} $\bx_0^{CF}$
    \vspace{.04in}
  \end{algorithmic}
\end{algorithm}

\subsection{Counterfactual Generation}
A fundamental property of causal models is the ability to perform interventions and observe changes to a system. In generative models, this enables the sampling of counterfactual data. Given a pre-trained CausalDiffAE, we can controllably manipulate any factor of variation, propagate the causal effects to descendants, and perform reverse diffusion to sample from the counterfactual distribution. Algorithm \ref{alg:generation} shows the process of generating counterfactuals from a trained CausalDiffAE, where $\bx_0$ refers to the factual observation and $\bx_0^{CF}$ refers to the generated counterfactual sample. To generate counterfactual instances, we first encode the high dimensional observation $\bx_0$ to a noise encoding $\bu$ (abduction) and transform it to causal latent variables $\zc$. Then, we intervene on a desired variable and propagate the causal effects via neural mechanisms to yield the intervened representation $\bar{\bz}_{\text{causal}}$. We utilize the DDIM sampling algorithm to ensure the stochastic noise $\bx_T$ is a deterministic encoding to enable semantic manipulations. Finally, we decode using DDIM conditioned on $(\bar{\bz}_{\text{causal}}, \bx_T)$ to obtain a counterfactual $\bx_0^{CF}$. In lines 12-13, we use the DDIM non-Markovian deterministic generative process to generate counterfactual instances as follows:

\begin{equation}
\begin{split}
    \bx_{t-1}^{CF} &= \sqrt{\alpha_{t-1}}\Big(\frac{\bx_t^{CF} - \sqrt{1-\alpha_t}\epsilon_{\theta}(\bx_t^{CF}, t, \bar{\bz}_{\text{causal}})}{\sqrt{\alpha_t}}\Big) \\&+ \sqrt{1 - \alpha_{t-1}}\epsilon_{\theta}(\bx_t^{CF}, t,\bar{\bz}_{\text{causal}})
\end{split}
\end{equation}

\textbf{Conditioning vs. Intervening.} When we study causal generative models, we utilize the intervention operation, which is a fundamentally different operation than conditioning. When we condition, we narrow our scope to a specific subgroup of the data based on the conditioning variable. Interventions are population-level operations that fix a variable's value (rendering it independent of its parents) to determine causal effects downstream. We emphasize that, under this intervention operation, causal models are robust to distribution shifts and can generate data outside the support of the training distribution. 

\subsection{Weak Supervision}
 To reduce the reliance on labeled data, inspired by classifier-free \citep{ho2021classifierfree} guidance, we train a CausalDiffAE with a weak supervision guidance paradigm on the representation level \citep{dhariwal2021diffusion}. 

\textbf{Training.} In the limited labeled-data regime, we train two models: an unconditional denoising diffusion model $p_{\theta}(\bx)$ parameterized by the score estimator $\epsilon_{\theta}(\bx_t, t)$ and a representation-conditioned model $p_{\theta}(\bx | \bz_{\text{causal}})$ parameterized through $\epsilon_{\theta}(\bx_t, t, \bz_{\text{causal}})$. We use a single neural network to parameterize both models, where for the unconditional model we use only the unlabeled data for predicting the score (i.e., $\epsilon_{\theta}(\bx_t, t)$).

\textbf{Generation.} The counterfactual generation procedure in lines 12-13 of Algorithm \ref{alg:generation} can be modified to generate counterfactuals with a guidance strength $\omega$, which can be interpreted as controlling the strength of the intervention on the causal variable to generate the counterfactual in our case. The overall modified score estimation during generation can be performed using the following linear combination of conditional and unconditional score estimates

\begin{equation}
    \bar{\epsilon}_{\theta}(\bx_t, t, \bar{\bz}_{\text{causal}}) = \underbrace{\omega\epsilon_{\theta}(\bx_t, t, \bar{\bz}_{\text{causal}})}_{\text{causal conditional model}} + \underbrace{(1-\omega)\epsilon_{\theta}(\bx_t, t)}_{\text{unconditional model}}
\end{equation}

where $\bar{\bz}_{\text{causal}}$ is the set of latent causal factors after an intervention. The original utility of the classifier-free paradigm was to decrease the generation of diverse data in favor of higher-quality image samples without needing classifier gradients. So, $\omega$ controls the trade-off between higher quality and diverse samples. In our case, we care about generating high-quality counterfactual data. Intuitively, a higher $\omega$ implies a stronger effect of the intervention on the generated counterfactual since the conditional model $\epsilon_{\theta}(\bx_t, t, \zc)$ is sensitive to interventions. So, as $\omega$ decreases, the unconditional model dilutes the effect of the intervention-sensitive model. In this sense, the sampling mechanism can be used to evaluate the causal strength of interventions. We find that the weak supervision paradigm enables (1) more efficient training with a weaker supervision signal, and (2) fine-grained control over generated counterfactuals.

\section{Experiments}

\subsection{Empirical Setting}

\begin{figure*}[t!]
    \centering
    \begin{subfigure}[t]{0.46\textwidth}
        \includegraphics[scale=0.47]{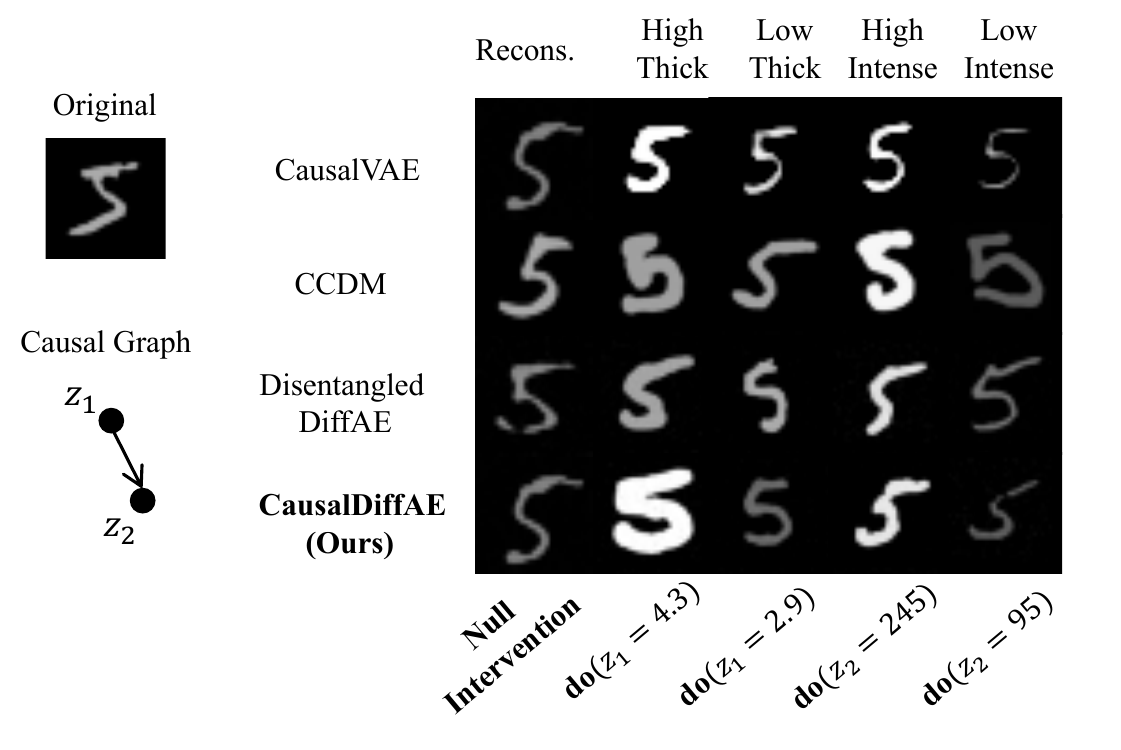}
        \caption{MorphoMNIST results (Orig: $y_1=2.399, y_2=162.2739$)}
        \label{fig:mmnist}
     \end{subfigure}
     ~
     \begin{subfigure}[t]{0.52\textwidth}
        \includegraphics[scale=0.47]{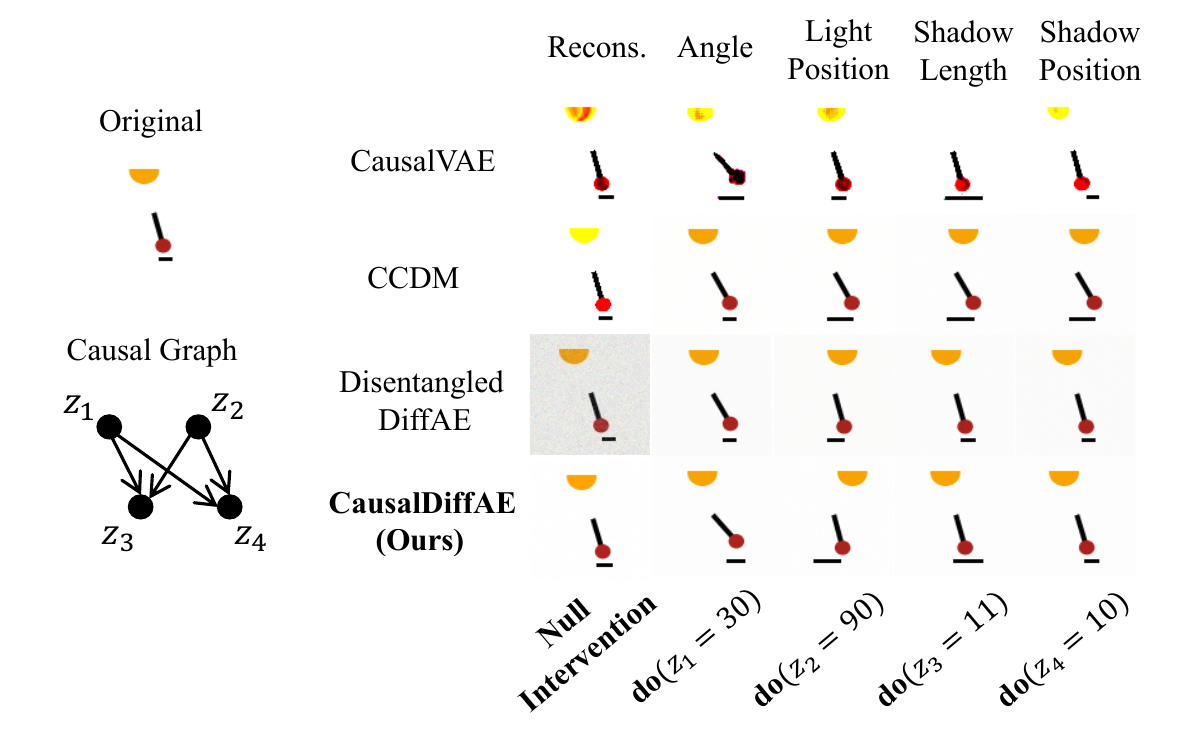}
        \caption{Pendulum results (Orig: $y_1 = 16, y_2 = 113, y_3 = 3, y_4 = 12$)}
        \label{fig:pend}
     \end{subfigure}
     \vspace{0.5cm}
     \caption{Counterfactual trajectories generated by CausalDiffAE and baseline models for (a) MorphoMNIST and (b) Pendulum datasets. We observe that CausalDiffAE generates much more accurate counterfactuals upon interventions on causal factors compared to baselines.}
     \vspace{0.2cm}
\end{figure*}

\textbf{Datasets.} We experiment on three datasets. We use the MorphoMNIST dataset \citep{castro2019morphomnist} created by imposing a 2-variable SCM to generate morphological transformations on the original MNIST dataset, where thickness is the cause of the intensity of the digit \citep{deepscm}, as shown in Figure \ref{fig:mmnist}. The Pendulum dataset \citep{DBLP:conf/cvpr/YangLCSHW21} consists of images of a causal system consisting of a light source, pendulum, and shadow. The light source and the pendulum angle determine the length and position of the shadow, as shown in the causal graph in Figure \ref{fig:pend}. We also use CausalCircuit \citep{brehmer2022weakly}, a complex 3D robotics dataset where a robot arm moves around to turn on red, green, or blue lights. The causal graph of this system is shown in Figure \ref{fig:circuit_counterfactuals}.

\textbf{Baselines.} CausalVAE \citep{DBLP:conf/cvpr/YangLCSHW21} is a VAE-based causal representation learning framework that models causal variables using a linear SCM and enables counterfactual generation during inference time through interventions on causal variables. Class-conditional diffusion model (CCDM) \citep{improved_diffusion} is a conditional diffusion model that utilizes class labels as the conditioning signal in reverse diffusion. Thus, this model is capable of generating new samples determined by a discrete or continuous set of labels $\by$. DiffAE
\citep{preechakul2021diffusion} is a diffusion model that aims to learn manipulable and semantically meaningful latent codes. However, this approach learns an arbitrary representation in an unsupervised fashion and does not disentangle the latent space. Manipulations are performed using a post-hoc classifier for linear interpolation. Thus, the learned representation would not be ideal to perform causal interventions. For a fair comparison in counterfactual generation, we modify the objective to disentangle the latent space by incorporating label information in a prior to regularize the posterior. We call this extension \textbf{DisDiffAE}. We use DisDiffAE as a baseline to evaluate counterfactual generation and the DiffAE to evaluate disentanglement.

\textbf{Metrics.} We primarily use two quantitative metrics to evaluate the performance of our approach. To evaluate the disentanglement of the learned representations, we use the \textit{DCI disentanglement} metric \citep{eastwood2023dcies}. A high DCI score also suggests the effectiveness of controllable generation. In the context of a causal representation, this means that we can intervene on latent codes in an isolated fashion without any entanglements (i.e., two factors are encoded in the same latent code). To quantitatively evaluate generated counterfactuals, we adopt the \textit{Effectiveness} metric from Melistas et al \citep{melistas2024benchmarking}, which evaluates how successful the performed intervention was at generating the counterfactual. We train anti-causal predictors via convolutional regressors on the training dataset for each continuous causal variable $z_i$. Then, we report the mean absolute error (MAE) loss between the predicted values from the generated counterfactual and the true values of the counterfactual. This metric captures how controllable the factors are and the accuracy of the generated counterfactuals. 

For details about the datasets, implementation, metrics, and computational requirements, see Appendix \ref{sec:exp_dets}. Our code is available at \url{https://github.com/Akomand/CausalDiffAE}.

\begin{table}[t]
\vspace{1cm}
\vskip-13pt
    \centering
    \footnotesize
    \setlength{\tabcolsep}{5pt}
        \caption{Disentanglement (DCI)}
        \vspace{0.5cm}
    \vskip-4pt
    \begin{tabular}{llcc}
        \toprule
        Dataset & Model & DCI $\uparrow$   \\
        \midrule
         MorphoMNIST & CausalVAE    &   $0.784\pm{0.01}$   \\ 
         & DiffAE  &  $0.358\pm{0.01}$     \\ \rowcolor{ourmethod} & CausalDiffAE  &   $\mathbf{0.993\pm{0.01}}$   \\ 
        \midrule
         Pendulum & CausalVAE   &   $0.885\pm{0.01}$    \\ 
         & DiffAE  &   $0.353\pm{0.01}$  \\ \rowcolor{ourmethod} & CausalDiffAE  &   $\mathbf{0.999\pm{0.01}}$  \\
        \midrule
         CausalCircuit & CausalVAE    &  $0.8860\pm{0.01}$  \\ 
         & DiffAE  &  $0.353\pm{0.01}$  \\ \rowcolor{ourmethod} & CausalDiffAE   &  $\mathbf{0.999\pm{0.01}}$  \\
        \bottomrule
    \end{tabular}
        \label{tab:full_supervision_disentanglement}
\end{table}

\subsection{Disentanglement of Latent Space}
We compare the disentanglement of CausalDiffAE with other baseline models, as shown in Table \ref{tab:full_supervision_disentanglement}. We observe that diffusion-based representation learning objectives coupled with a suitable prior can better disentangle latent variables compared to VAE-based models. We do not include CCDM as a baseline here since it does not produce a representation to be evaluated. Compared to CausalVAE, the diffusion-based decoder in CausalDiffAE disentangles the semantic factors of variation to a much greater degree. Thus, we can perform interventions on causal variables in isolation and observe their downstream effects. We also note that DiffAE \citep{preechakul2021diffusion} does not learn a disentangled latent space since the semantic representation learned is arbitrary. To perform controllable manipulations with DiffAE, a post-hoc classifier must be trained to guide the sampling process. CausalDiffAE offers more precise control over learned factors through the disentanglement objective without the need to train additional classifiers.

\subsection{Controllable Counterfactual Generation}

\textbf{Qualitative Evaluation.} We show that CausalDiffAE produces much more realistic counterfactual samples compared to other acausal baselines and its VAE counterpart, CausalVAE. We attribute this to the diffusion process, which is better capable of capturing causally relevant information along with low-level stochastic variation.

Figure \ref{fig:mmnist} shows the counterfactual generation results for the MorphoMNIST dataset. CausalVAE can generate counterfactual images after intervening on either thickness or intensity, but the accuracy and quality of the generated counterfactuals is far lower than CausalDiffAE. For instance, lower thickness does not lower the intensity and lower intensity intervention seems to change the thickness of the digit. CCDM fails to produce samples consistent with the underlying causal model. For example, intervening on the intensity produces a sample that increases in thickness. From a conditioning perspective, high-intensity digits tend to be thicker in the training distribution. For DisDiffAE, increasing the thickness does not influence the intensity. 

\begin{figure}[t]
    \centering
        \includegraphics[width=0.5\textwidth]{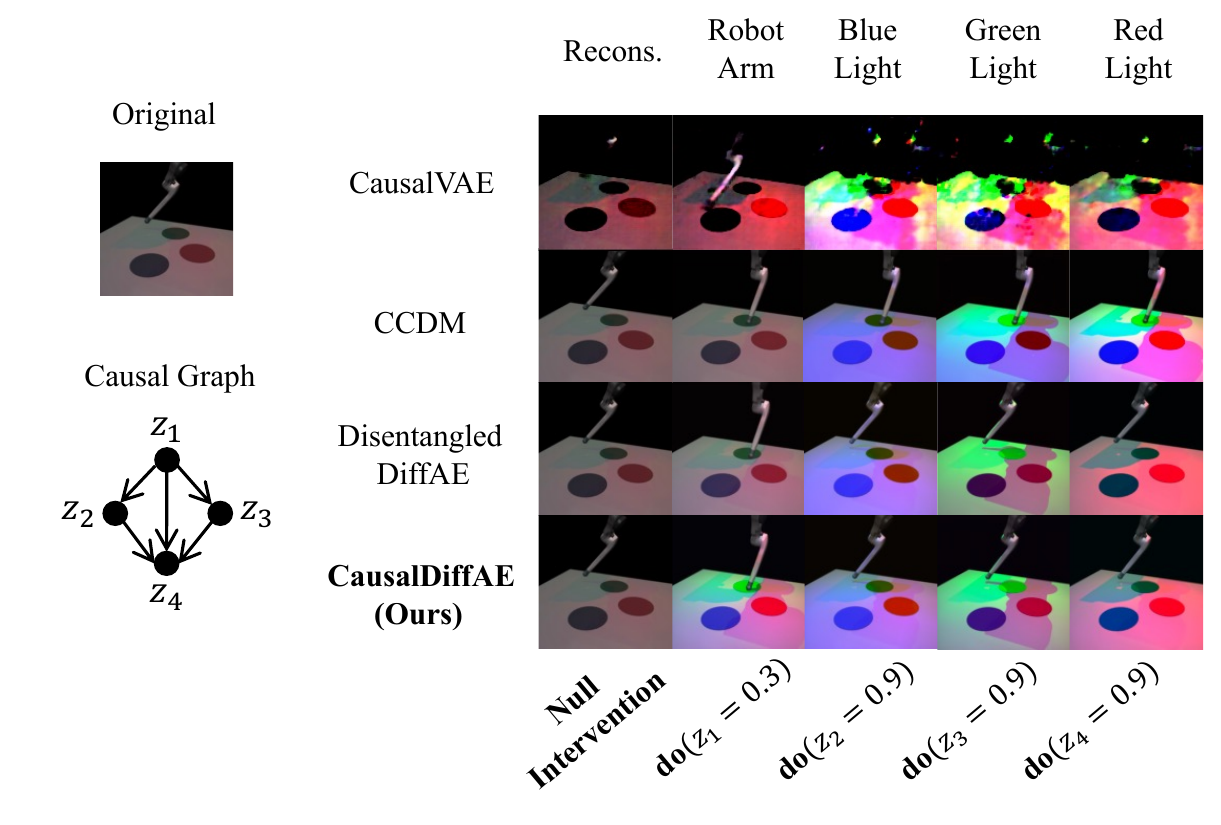}
        \caption{CausalCircuit results (Orig: $y_1 = 0.02, y_2 = 0.03, y_3 = 0.04, y_4 = 0.14$)}
        \vspace{0.5cm}
    \label{fig:circuit_counterfactuals}
\end{figure}

Figure \ref{fig:pend} shows counterfactual generation results for the Pendulum causal system. Upon interventions, images generated by CCDM are not consistent with the causal model. For example, intervening on the light position changes both the light position and the pendulum angle. DisDiffAE produces images where we can control one factor at a time, but does not reflect causal effects. For example, changing the angle of the pendulum does not accurately change the shadow length and position. CausalDiffAE generates higher-quality counterfactuals that are consistent with the causal model. Specifically, intervention on the pendulum angle or light position changes the shadow length and position accurately. On the other hand, interventions on children variables leave the parents unchanged.

Figure \ref{fig:circuit_counterfactuals} shows counterfactuals from the CausalCircuit dataset. CausalVAE generates inaccurate counterfactuals for many scenarios (e.g., intervention on the blue light intensity changes the intensity of all other lights). For CCDM, moving the robot arm over the green light fails to turn the light on. Furthermore, manipulating the light intensity of other lights affects the position of the robot arm. DisDiffAE enables control over the generative factors, but does not consider causal effects. For example, moving the robot arm over the green light does not turn it on. Counterfactuals generated from CausalDiffAE are consistent with the causal system. For example, moving the robot arm over the green button turns the light on and as a result also turns on the red light, which is a downstream child variable. Intervening on the blue or green light slightly increases the intensity of the red light. Intervening on the red light leaves all parent variables unchanged. For additional counterfactual generation results, see Appendix \ref{sec:additional_exp}.

\begin{table}[t]
\vspace{0.5cm}
\vskip-13pt
    \centering
    \footnotesize
    \setlength{\tabcolsep}{5.5pt}
        \caption{Effectiveness on MorphoMNIST test set (MAE)}
        \vspace{0.5cm}
    \vskip-8pt
    \begin{tabular}{cccc}
        \toprule
        \multirow{2}{*}{Factor} & \multirow{2}{*}{Model} & \multicolumn{2}{c}{Intervention} \\
        \cmidrule{3-4}
         &  & \textbf{do}($t$) & \textbf{do}($i$) \\
        \midrule
        Thickness & CausalVAE  & $3.763\pm{0.01}$ & $4.645\pm{0.01}$ \\
         ($t$) & DisDiffAE & $\mathbf{0.377\pm{0.02}}$ & $0.326\pm{0.02}$  \\ \rowcolor{ourmethod} & CausalDiffAE & $0.392\pm{0.02}$ & $\mathbf{0.309\pm{0.02}}$ \\
        \midrule
        Intensity & CausalVAE & $13.233\pm{0.01}$ & $15.087\pm{0.01}$ \\
         ($i$) & DisDiffAE & $0.794\pm{0.02}$ & $0.262\pm{0.02}$ \\ \rowcolor{ourmethod} & CausalDiffAE  & $\mathbf{0.503\pm{0.01}}$ & $\mathbf{0.256\pm{0.01}}$ \\
        \bottomrule
    \end{tabular}
        \label{tab:effectiveness_morpho}
\end{table}

\begin{table}[t]
\vspace{0.5cm}
\vskip-13pt
    \centering
    \footnotesize
    \setlength{\tabcolsep}{5.5pt}
        \caption{Effectiveness on Pendulum test set (MAE)}
        \vspace{0.5cm}
    \vskip-8pt
    \begin{tabular}{cccccc}
        \toprule
        \multirow{2}{*}{Factor} & \multirow{2}{*}{Model} & \multicolumn{4}{c}{Intervention} \\
        \cmidrule{3-6}
         &  & \textbf{do}($a$) & \textbf{do}($lp$) & \textbf{do}($sl$) & \textbf{do}($sp$) \\
        \midrule
        Angle & CausalVAE  & $24.860$ & $23.030$ & $20.470$ & $11.580$ \\
         ($a$) & DisDiffAE & $0.668$ & $0.648$ & $0.647$ & $0.647$  \\ \rowcolor{ourmethod} & CausalDiffAE & $\mathbf{0.297}$ & $\mathbf{0.132}$ & $\mathbf{0.031}$ & $\mathbf{0.034}$ \\
        \midrule
        LightPos & CausalVAE & $34.200$ & $26.010$ & $35.490$ & $47.060$ \\
         ($lp$) & DisDiffAE & $0.656$ & $0.654$ & $0.630$ & $0.651$ \\ \rowcolor{ourmethod} & CausalDiffAE  & $\mathbf{0.045}$ & $\mathbf{0.434}$ & $\mathbf{0.035}$ & $\mathbf{0.064}$ \\
        \midrule
        ShadowLen & CausalVAE & $1.946$ & $1.43$ & $2.02$ & $1.72$ \\
         ($sl$) & DisDiffAE & $0.550$ & $0.527$ & $0.560$ & $0.516$ \\ \rowcolor{ourmethod} & CausalDiffAE & $\mathbf{0.136}$ & $\mathbf{0.322}$ & $\mathbf{0.492}$ & $\mathbf{0.082}$ \\
        \midrule
        ShadowPos & CausalVAE & $52.52$ & $72.50$ & $57.03$ & $32.78$ \\
        ($sp$) & DisDiffAE & $0.474$ & $0.475$ & $0.479$ & $0.534$ \\ \rowcolor{ourmethod} & CausalDiffAE & $\mathbf{0.146}$ & $\mathbf{0.303}$ & $\mathbf{0.064}$ & $\mathbf{0.471}$ \\
        \bottomrule
    \end{tabular}
        \label{tab:effectiveness_pend}\\ 
        * Standard error is roughly in the range $\pm0.01$ to $\pm0.02$ for all averages.
\end{table}

\textbf{Quantitative Evaluation.} We quantitatively show using the effectiveness metric that CausalDiffAE generates counterfactuals that are both accurate and realistic. We perform random interventions from a uniform distribution over the test dataset for each causal variable. We find that CausalDiffAE almost always outperforms other baselines in the effectiveness metric, as shown in Tables \ref{tab:effectiveness_morpho} and \ref{tab:effectiveness_pend}, for all causal factors. Specifically, for the MorphoMNIST dataset, we observe that interventions on thickness produce counterfactuals that accurately reflect both the thickness and intensity values. In the scenario where we intervene on thickness, the intensity MAE is lower for CausalDiffAE than other baselines, which indicates that the generated counterfactual has an accurate intensity value consistent with the causal effect of thickness on intensity. When we intervene on intensity, the thickness MAE is lower for CausalDiffAE than baselines, which suggests that the generated counterfactual retains its original thickness value upon intervention on intensity. For the Pendulum dataset, we see a similar phenomenon, where interventions on causal factors along with their downstream effects are accurately captured in the generated counterfactuals. We do not evaluate effectiveness for the CausalCircuit dataset since we do not have access to the generative process used to obtain the factors. We compute the average effectiveness value over $5$ runs with different random seeds. Our results strongly imply that the generated counterfactuals closely match the true counterfactuals. 

\subsection{Case Study: Weak Supervision Results}
\begin{figure}[t]
        \includegraphics[scale=1]{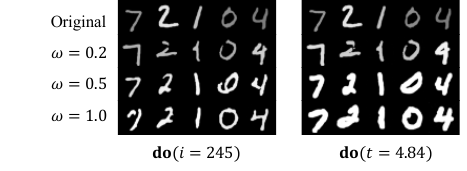}
        \caption{MorphoMNIST Weak Supervision}
        \vspace{0.5cm}
    \label{fig:morphomnist_reduced}
\end{figure}

Unlike VAE-based approaches, the weak supervision paradigm of diffusion models reduces the full-label supervision. We study the weak supervision scenario with the MorphoMNIST dataset. We jointly train a representation-conditioned and unconditional model, where the conditioned split is far less than the unconditioned split. We have two main motivations for doing this: (1) it greatly reduces the need for fully labeled datasets, and (2) it enables granular control over generated counterfactuals. We denote the proportion of unlabeled data by $p_{\text{unlabeled}}$. The CausalDiffAE model trained with $p_{\text{unlabeled}} = 0.8$ on the MorphoMNIST dataset yields a DCI score of $0.9964$, which suggests that even under strictly limited label supervision, CausalDiffAE learns disentangled representations. We also empirically show that changing the $\omega$ parameter controls the strength of the intervention on the generated counterfactual. Figure \ref{fig:morphomnist_reduced} shows MNIST digits generated using the joint estimated score from the reduced supervision version of CausalDiffAE. We observe that interventions have virtually no effect when sampling using the joint score with $\omega=0.2$. For $\omega=0.5$, we see a stronger effect of the intervention on the thickness and intensity of the digit. Finally, for the fully-supervised score $\omega=1.0$, the intervention acts the strongest. Thus, varying $\omega$ in the range $(0, 1)$ can be interpreted as generating a range of different counterfactuals.

\section{Conclusion}
In this work, we propose CausalDiffAE, a diffusion-based framework for causal representation learning and counterfactual generation. We propose a causal encoding mechanism that maps images to causally related factors. We learn the causal mechanisms among factors via neural networks. We formulate a variational diffusion-based objective to enforce the disentanglement of the latent space to enable latent space manipulations. We propose a DDIM-based counterfactual generation algorithm subject to interventions. For limited supervision scenarios, we propose a weak supervision extension of our model, which jointly learns an unconditional and conditional model. This objective also enables granular control over generated counterfactuals. We empirically show the capability of our model using both qualitative and quantitative metrics. Future work includes exploring counterfactual generation in text-to-image diffusion models. 

\section*{Acknowledgements}

This work is supported in part by National Science Foundation under awards 1910284, 1946391 and 2147375, the National Institute of General Medical Sciences of National Institutes of Health under award P20GM139768, and the Arkansas Integrative Metabolic Research Center at University of Arkansas.

\bibliography{main}
\onecolumn
\appendix
\section*{Appendices}

\section{Derivation of ELBO}
\label{sec:elbo}
Given a high-dimensional input image $\bx_0$, an auxiliary weak supervision signal $\by$, a latent noise encoding $\bu$, latent representation $\bz_{\text{causal}}$, and a sequence of $T$ latent representations $\bx_{1:T}$ learned by the diffusion model, the CausalDiffAE generative process can be factorized as follows:

\begin{equation}
    p(\bx_{0:T}, \bu, \bz_{\text{causal}} | \by) = p_{\theta}(\bx_{0:T} | \bu, \zc, \by)p(\bu, \zc|\by)
\end{equation}
where $\theta$ are the parameters of the reverse process of the conditional diffusion model. The log-likelihood of the input data distribution can be obtained as follows:
\begin{equation}
    \log p(\bx_0, \by) = \log \int p(\bx_{0:T}, \bu, \zc, \by) \; d\bx_{1:T}\;d\bu\;d\zc
\end{equation}
The joint posterior distribution $p(\bx_{1:T}, \bu, \zc | \bx_0, \by)$ is intractable, so we approximate it using a variational distribution $q(\bx_{1:T}, \bu, \zc | \bx_0, \by)$ which can be factorized into the following conditional distributions
\begin{equation}
    q(\bx_{1:T}, \bu, \zc | \bx_0, \by) = q_{\phi}(\zc, \bu | \bx_0, \by)q(\bx_{1:T}|\bu, \zc, \bx_0)
\end{equation}
where $\phi$ are the parameters of the variational encoder network. Since the likelihood of the data is intractable, we can approximate it by maximizing the following evidence lower bound (ELBO):

\begin{align}
    \log p(\bx_0, \by) &\geq \mathbb{E}_{q(\bx_{1:T}, \bu, \zc | \bx_0, \by)}\Bigg[\log \frac{p(\bx_{0:T}, \bu, \zc, \by)}{q(\bx_{1:T}, \bu, \zc | \bx_0, \by)}\Bigg] \\
    &=  \mathbb{E}_{q(\bx_{1:T}, \bu, \zc | \bx_0, \by)} \Bigg[\log \frac{p(\bu)p(\zc|\by)p_{\theta}(\bx_{0:T} | \bu, \zc)}{q_{\phi}(\zc, \bu | \bx_0, \by)q(\bx_{1:T}|\bu, \zc, \bx_0)}\Bigg] \\
    &= \mathbb{E}_{q(\bx_{1:T}, \bu, \zc | \bx_0, \by)} \Bigg[\log \frac{p(\bu, \zc | \by)}{q_{\phi}(\zc, \bu | \bx_0, \by)} + \log \frac{p_{\theta}(\bx_{0:T} | \bu, \zc)}{q(\bx_{1:T}|\bu, \zc, \bx_0)}\Bigg] \\
    &= \mathbb{E}_{q(\bu, \zc | \bx_0, \by)} \Bigg[\log \frac{p(\bu, \zc | \by)}{q_{\phi}(\zc, \bu | \bx_0, \by)}\Bigg] + \mathbb{E}_{q(\bx_{1:T}, \bu, \zc | \bx_0)} \Bigg[\log \frac{p_{\theta}(\bx_{0:T} | \bu, \zc)}{q(\bx_{1:T}|\bu, \zc, \bx_0)}\Bigg] \\
    &= \mathbb{E}_{q(\bu, \zc | \bx_0, \by)}\Bigg[\underbrace{\mathbb{E}_{q(\bx_{1:T}, \bu, \zc | \bx_0)} \Bigg[\frac{p_{\theta}(\bx_{0:T} | \bu, \zc)}{q(\bx_{1:T}|\bu, \zc, \bx_0)}\Bigg]}_{\text{Representation-conditioned DDPM Loss}}\Bigg] - \underbrace{\mathcal{D}_{KL}(q_{\phi}(\bu, \zc | \bx_0, \by) \| p(\bu, \zc | \by))}_{\text{Joint Latent Posterior Loss}} \\
\end{align}

In the learning process, we minimize the negative of the derived ELBO. We simplify this objective by using the $\epsilon_{\theta}$ parameterization to optimize the representation-conditioned DDPM loss. Further, since $\bu$ and $\zc$ are one-to-one mapped, we can split the joint conditional distribution into separate conditional distributions. Thus, we have the following final objective for CausalDiffAE:

\begin{equation}
    \mathcal{L}_{\text{CausalDiffAE}} = \sum_{t=1}^T \mathbb{E}_{t, \bx_0, \epsilon} \Big[ \|\epsilon_{\theta}(\bx_t, t, \bz_{\text{causal}}) - \epsilon_t \|_2^2\Big] + \gamma \Big\{ \mathcal{D}_{KL}(q_{\phi}(\bz_{\text{causal}} | \bx_0, \by) \| p(\bz_{\text{causal}} | \by)) + \mathcal{D}_{KL}(q_{\phi}(\bu | \bx_0) \| \mathcal{N}(\mathbf{0}, \mathbf{I}))\Big\}
\end{equation}

\section{Connection to Score-based Generative Models}
\label{sec:score_connection}
Diffusion models can also be represented as stochastic differential equations (SDEs) \citep{song2021scorebased} to model continuous-time perturbations. Specifically, the forward diffusion process can be modeled as the solution to an SDE on a continuous-time domain $t\in [0, T]$ with stochastic trajectories:
\begin{equation}
    d\bx = f(\bx, t) \;dt + g(t) \;dw
    \label{eq:forwardsde}
\end{equation}
where $w$ is the standard Weiner process, $f$ is a vector-valued function known as the drift coefficient of $\bx(t)$ and $g$ is a scalar function known as the diffusion coefficient of $\bx(t)$. The drift and diffusion coefficients can be considered as the mean and variance of the noise perturbations in the diffusion process, respectively. The reverse diffusion process can be modeled by the solution to the reverse-time SDE of Eq. (\ref{eq:forwardsde}), which can be derived analytically as:
\begin{equation}
    d\bx = [f(\bx, t) - g^2(t)\nabla_x \log p_t(\bx)] \;dt + g(t) \;d\bar{w}
    \label{eq:reversesde}
\end{equation}
where $\bar{w}$ is the standard Weiner process in reverse time and $\nabla_x \log p_t(\bx)$ is the score of the data distribution at timestep $t$. Once we know the score of the marginal distribution for all timesteps $t$, we can derive the reverse diffusion process from Eq. (\ref{eq:reversesde}).

Song et al \citep{song2021scorebased} showed that the denoising diffusion probabilistic model (DDPM) is a discretization of the following Variance Preserving SDE (VP-SDE)
\begin{equation}
    d\bx = \frac{1}{2}\beta(t)\bx \; dt + \sqrt{\beta(t)} \; dw
\end{equation}
Thus, learning a noise prediction network $\epsilon_{\theta}$ and minimizing MSE in diffusion probabilistic models is equivalent to approximating the score of the data distribution in the SDE formulation. 
 From a score-based perspective, we aim to minimize the following conditional denoising score-matching form of our objective
\begin{equation}
\begin{split}
     &\mathbb{E}_{p(\bx)} \mathbb{E}_{q_{\phi}(\bz_{\text{causal}}|\bx_0)}\mathbb{E}_{q(\bx_t | \bx_0)} \Big[\log p(\bu) + p(\bz_{\text{causal}} | \by) \\& - \log q_{\phi}(\bu | \bx_0) - \log q_{\phi}(\bz_{\text{causal}} | \bx_0, \by) \\& + \lambda(t) \|s_{\theta}(\bx_t, \bz_{\text{causal}}, t) - \nabla_{\bx_t} \log p(\bx_t | \bx_0) \| \Big]
\end{split}
\end{equation}
where $s_{\theta}$ approximates the score of the data distribution conditioned on $\bx_0$ and $\lambda(t)$ is a positive weighing function. The ideal for modeling natural phenomena in the world is by using differential equations to model the physical mechanisms \citep{scholkopf_toward_2021}. In the SDE formulation, the causal variables are used to denoise the high-dimensional data, which is modeled as a reverse-time stochastic trajectory. We can interpret this idea as modeling the dynamics of high-dimensional systems by incorporating causal information. As opposed to simply learning an arbitrary latent representation, a disentangled causal representation encodes the causal information that the denoising process can use to reconstruct \textit{causally relevant} features in high-dimensional data.

\section{Discussion on Causal Discovery}
\label{sec:limitations}
In this work, we assume the latent causal structure is known since we focus on counterfactual generation. In principle, our framework can be combined with causal structure learning methods such as NOTEARS \cite{NOTEARS} by adding a penalty to terms in the VAE loss objective to enforce sparsity and acyclicity as follows
\begin{equation}
    \mathcal{L}_{total} = \mathcal{L}_{\text{CausalDiffAE}} + H(A) + \|A\|_0
\end{equation}
where $H(A) = tr[(I + \alpha A \odot A)]^n - n = 0$ is the acyclicity constraint and $\|\cdot\|_0$ enforces the sparsity of the DAG. We can alternatively use the $\|\cdot\|_1$ for sparsity to ensure a differentiable objective. Similar to \cite{NOTEARS}, we can utilize the augmented Lagrangian to optimize the joint loss objective. Additionally, other causal discovery algorithms could be used heuristically with a variety of different assumptions \cite{vowels2021d}. We look to explore this direction in future work.

\section{Experiment Details}
\label{sec:exp_dets}
\subsection{Dataset Details}
\label{sec:datasets}
\textbf{MorphoMNIST.} The MorphoMNIST dataset \cite{castro2019morphomnist} is produced by applying morphological transformations on the original MNIST handwritten digit dataset. The digits can be described by measurable shape attributes such as stroke thickness, stroke length, width, height, and slant of digit. Pawlowski et al \cite{deepscm} impose a $3$-variable SCM to generate the morphological transformations, where stroke thickness is a cause of the brightness of each digit. That is, thicker digits are often brighter, whereas thinner digits are dimmer. The data-generating process is as follows

\begin{equation}\label{eq:true_scm}
\begin{aligned}
    t = &f_T(u_T) = 0.5 + u_T \,, &
    u_T &\sim \Gamma(10, 5) \,, \\
    i = &f_I(u_I; t) = 191 \cdot \sigma(0.5\cdot u_I + 2\cdot t - 5) + 64 \,, &
    u_I &\sim \mathcal{N}(0, 1) \,, \\
    x = &f_X(u_X; i, t) = \text{SetIntensity(SetThickness($u_X; t$) $; i$)} \,, &
    u_X &\sim \text{MNIST} \,,
\end{aligned}
\end{equation}
where $x$ is the resulting image, $u$ is the exogenous noise for each variable, and $\sigma(\cdot)$ is the logistic sigmoid.

\noindent
\textbf{Pendulum.} The Pendulum dataset \cite{DBLP:conf/cvpr/YangLCSHW21} consists of a set of $7$K images with resolution $96\times 96\times 4$ describing a physical system of a pendulum and light source that cause the length and position of a shadow. The causal variables of interest are the angle of the pendulum, the position of the light source, the length of the shadow, and the position of the shadow. The data generating process is as follows:

\begin{align*}
    y_1 &\sim U(-45, 45);\; \qquad \theta = y_1 * \frac{\pi}{200} ;\; \qquad x = 10 + 9.5\sin\theta \\
    y_2 &\sim U(60, 145);\; \qquad \phi = y_2 * \frac{\pi}{200} ;\; \qquad y = 10 - 9.5\cos\theta \\
    y_3 &= \max(3, \Big|9.5\frac{\cos\theta}{\tan\phi} + 9.5\sin\theta\Big|) \\
    y_4 &= \frac{-11 + 4.75\cos\theta}{\tan\phi} + (10 + 4.75\sin\theta)
\end{align*}

\noindent
\textbf{Causal Circuit}. The Causal Circuit dataset is a new dataset created by \cite{brehmer2022weakly} to explore research in causal representation learning. The dataset consists of $512\times 512\times 3$ resolution images generated by $4$ ground-truth latent causal variables: robot arm position, red light intensity, green light intensity, and blue light intensity. The images show a robot arm interacting with a system of buttons and lights. The data is rendered using an open-source physics engine. The original dataset consists of pairs of images before and after an intervention has taken place. For the purposes of this work, we only utilize observational data of either the before or after system. The data is generated according to the following process:
\begin{align*}
    v_R &= 0.2 + 0.6 * \text{clip}(y_2 + y_3 + b_R, 0, 1) \\
    v_G &= 0.2 + 0.6 * b_G \\
    v_B &= 0.2 + 0.6 * b_B \\
    y_4 &\sim \text{Beta}(5v_R, 5 * (1 - v_R)) \\
    y_3 &\sim \text{Beta}(5v_G, 5 * (1 - v_G)) \\
    y_2 &\sim \text{Beta}(5v_B, 5 * (1 - v_B)) \\
    y_1 &\sim U(0, 1)
\end{align*}

where $b_R$, $b_G$, and $b_B$ are the pressed state of buttons that depends on how far the button is touched from the center, $y_1$ is the robot arm position, and $y_2$, $y_3$, and $y_4$ are the intensities of the blue, green, and red lights, respectively.

\subsection{Implementation Details}
\label{sec:implementation}
We use the same network architectures and hyperparameters used in other works based on diffusion models \citep{ho_ddpm, dhariwal2021diffusion, improved_diffusion}. We set the causal latent variable size to $512$ to ensure a large enough capacity to capture causally relevant information. The representation-conditioned noise predictor is parameterized by a UNet with the attention mechanism. Similar to \citep{ho_ddpm}, we use a linear noise scheduling for the variance parameter $\beta$ between $\beta_1 = 10^{-4}$ and $\beta_2=0.02$ during training. For all three datasets, we start the bottleneck parameter at $\gamma=0$ and linearly increase $\gamma$ throughout training to a final value of $\gamma=1.0$.

\begin{table}[t]
\vskip-13pt
\small
    \centering
    \setlength{\tabcolsep}{20pt}
        \caption{Implementation details of CausalDiffAE}
    \vskip-2pt
    \begin{tabular}{lccc}
        \toprule
        Parameter & MorphoMNIST & Pendulum & CausalCircuit \\
        \midrule
        Batch size & $768$ & $128$ & $128$ \\
        Base channels & $128$ & $128$ & $128$ \\
        Channel multipliers & $[1, 2, 2]$ & $[1, 2, 4, 8]$ & $[1, 2, 4, 8]$ \\
        Training set & $60$K & $5$K & $50$K \\
        Test set & $10$K & $2$K & $10$K \\
        Image resolution & $28\times 28\times 1$ & $96 \times 96 \times 4$ & $128 \times 128 \times 3$ \\
        Num causal variables & $2$ & $4$ & $4$ \\
        $z_{\text{causal}}$ size & $512$ & $512$ & $512$ \\
        $\beta$ scheduler & Linear & Linear & Cosine \\
        Learning rate & $10^{-4}$ & $10^{-4}$ & $10^{-4}$ \\
        Optimizer & Adam & Adam & Adam \\
        Diffusion steps & $1000$ & $1000$ & $4000$ \\
        Iterations & $10$K & $40$K & $20$K \\
        Diffusion loss & MSE & MSE & MSE \\
        Sampling & DDIM & DDIM & DDIM \\ 
        Stride & $250$ & $250$ & $250$ \\ 
        Bottleneck $\gamma$ & $1.0$ & $1.0$ & $1.0$ \\
        \bottomrule
    \end{tabular}
        \label{tab:implementation}
\end{table}

\subsection{Metrics Details}
\textbf{DCI Disentanglement \citep{eastwood2018a}.} The DCI disentanglement score quantifies the degree to which a representation disentangles the underlying factors of variation with each variable capturing \textit{at most one} generative factor. Let $P_{ij} = R_{ij} / \sum_{k=0}^{K-1} R_{ik}$ be the probability of $z_i$ being a strong predictor of $y_j$. Then, the disentanglement score is defined as
    \begin{equation}
        D_i = (1 + \sum_{k=0}^{K-1}P_{ik} \log_kP_{ik})
    \end{equation}
    If $z_i$ is a strong predictor for only a single generative factor, $D_i = 1$. If $z_i$ is equally important in predicting all generative factors, $D_i=0$. Let $\rho_i = \sum_j R_{ij} / \sum_{ij} R_{ij}$ be the relative latent code importances. The total disentanglement score is a weighted average of the individual $D_i$ 
    \begin{equation}
        D = \sum_i \rho_iD_i
    \end{equation}

\noindent
\textbf{Effectiveness \citep{melistas2024benchmarking}.} The effectiveness metric aims to identify how successful the performed intervention is. To quantitatively evaluate the effectiveness for a given counterfactual image, an anti-causal predictor $h_{\theta}^i$ is trained on the data distribution, for each causal variable $y_i$. Each predictor approximates the counterfactual value of the variable $y^{i*}_x$ given the counterfactual image $x^*$ as input
\begin{equation}
    \text{effectiveness}_i(x^*, y^{i*}_x) = d(y^{i*}_x, h_{\theta}^i(x^*))
\end{equation}
where $d(\cdot)$ is the corresponding distance, defined as a classification metric for categorical variables and a regression metric for continuous ones.

\subsection{Computational Requirements}
 We run our experiments on an Ubuntu 20.04 workstation with eight NVIDIA Tesla V100-SXM2 GPUs with 32GB RAM. It is well-known that diffusion models have a higher computational complexity than other generative models, such as VAEs and GANs. Generally speaking, all the diffusion-based approaches have quite a similar runtime, whereas CausalVAE is much faster. We expect any developments in the training and sampling efficiency of diffusion probabilistic models to apply to our proposed diffusion-based approach as well.

\subsection{Additional Experiments}

\label{sec:additional_exp}
\begin{figure}[H]
    \centering
    \includegraphics[scale=0.6]{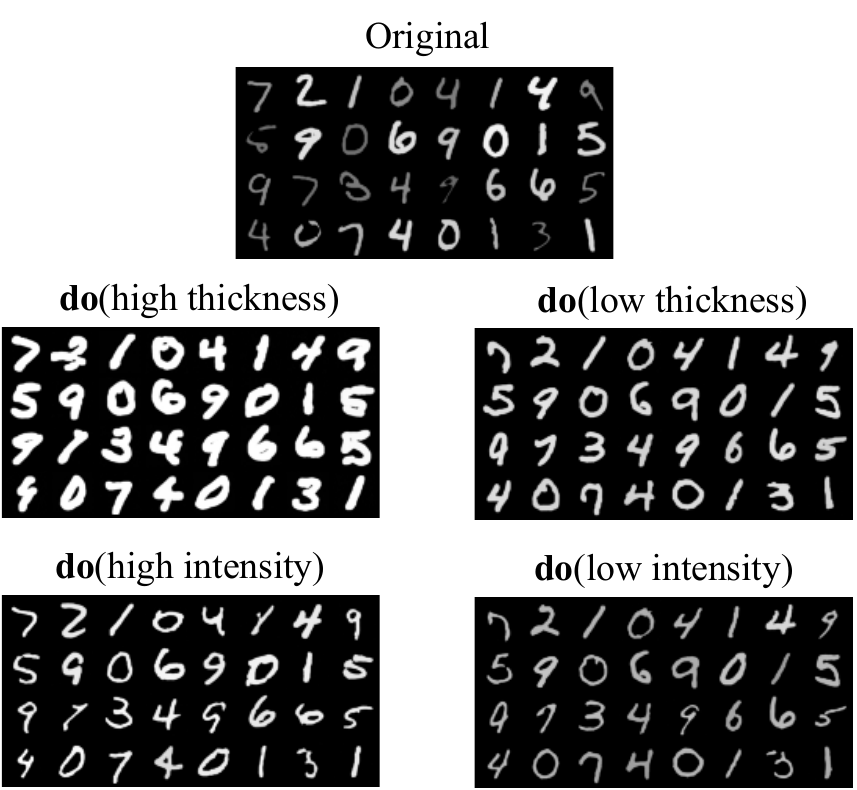}
    \caption{CausalDiffAE generated counterfactuals (MorphoMNIST)}
    \label{fig:appendix_mmnist_results}
\end{figure}

\begin{figure}[H]
    \centering
    \includegraphics[width=0.7\textwidth]{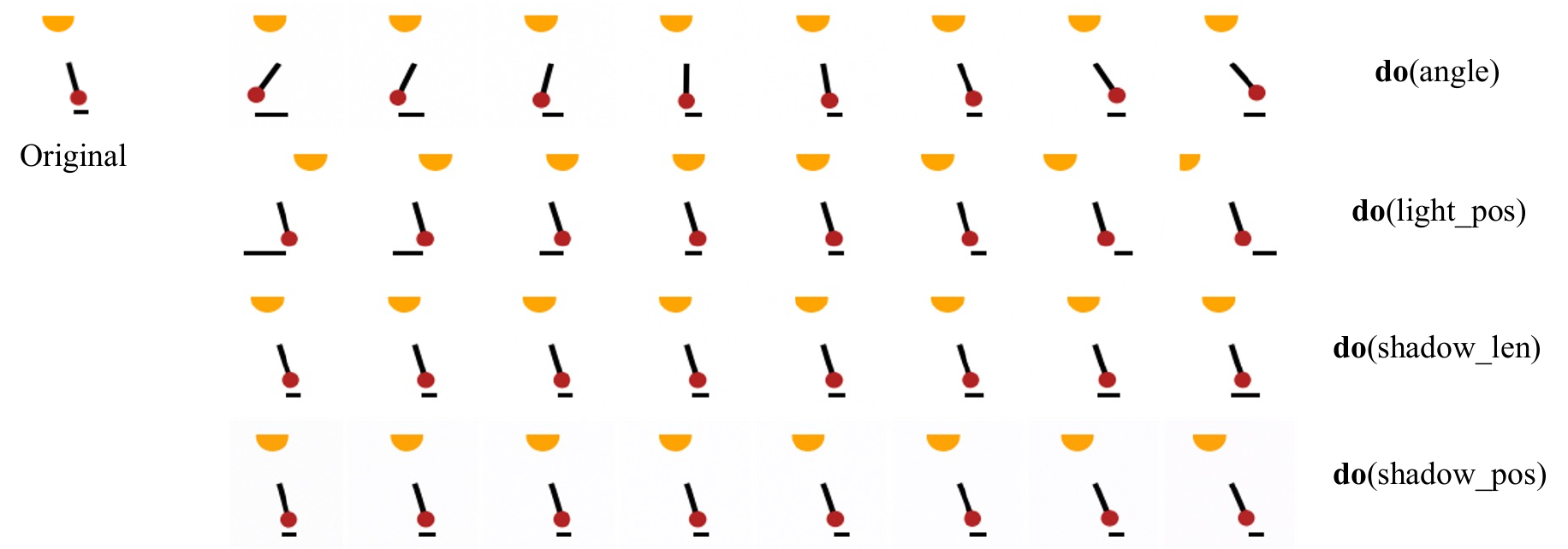}
    \caption{CausalDiffAE generated counterfactuals via latent traversals in the normalized range $(-1, 1)$ (Pendulum)}
    \label{fig:appendix_pendulum_results}
\end{figure}

\begin{figure}[H]
    \centering
    \includegraphics[scale=0.6]{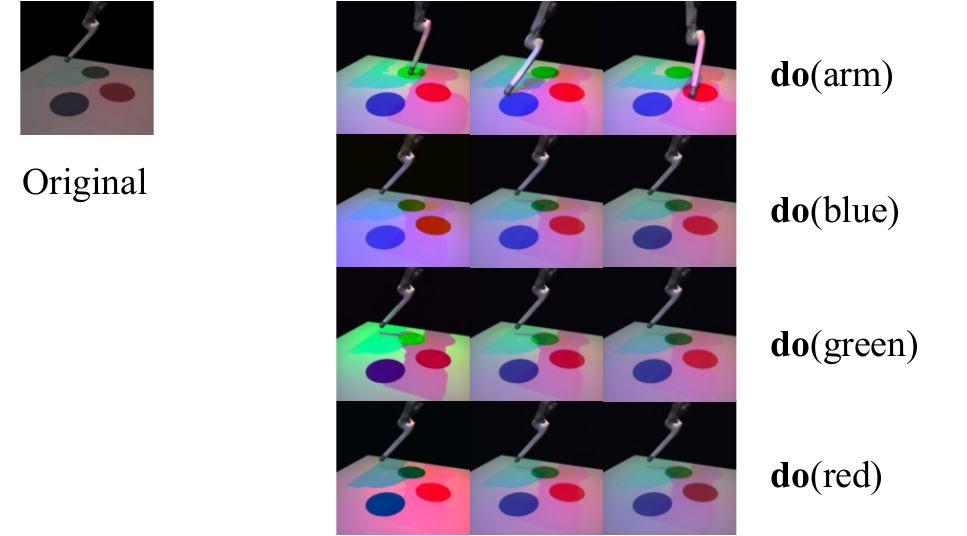}
    \caption{CausalDiffAE generated counterfactuals via latent traversals in the normalized range $(-1, 1)$ (CausalCircuit).}
    \label{fig:appendix_circuit_results}
\end{figure}

\end{document}